\documentclass{article}
\usepackage[T1]{fontenc} 
\usepackage[utf8]{inputenc} 
\usepackage{ismir,amsmath,cite,url}
\usepackage{graphicx}
\usepackage{booktabs}
\usepackage{color}
\usepackage{quoting}
\def\lbd{}	
\usepackage{booktabs}

\title{L\MakeLowercase{y}C\MakeLowercase{on}: Lyrics Reconstruction from the Bag-of-Words Using Large Language Models}

\twoauthors
  {Haven Kim} {University of California San Diego}
  {Kahyun Choi} {University of Illinois at Urbana-Champaign}

\def\authorname{Haven Kim and Kahyun Choi}

\begin{document}

\maketitle
\begin{abstract}
This paper addresses the unique challenge of conducting research in lyric studies, where direct use of lyrics is often restricted due to copyright concerns. Unlike typical data, internet-sourced lyrics are frequently protected under copyright law, necessitating alternative approaches. Our study introduces a novel method for generating copyright-free lyrics from publicly available Bag-of-Words (BoW) datasets, which contain the vocabulary of lyrics but not the lyrics themselves. Utilizing metadata associated with BoW datasets and large language models, we successfully reconstructed lyrics.  We have compiled and made available a dataset of reconstructed lyrics, LyCon, aligned with metadata from renowned sources including the Million Song Dataset, Deezer Mood Detection Dataset, and AllMusic Genre Dataset, available for public access. We believe that the integration of metadata such as mood annotations or genres enables a variety of academic experiments on lyrics, such as conditional lyric generation.

\end{abstract}

\section{Introduction}
One of the distinct challenges in lyric research is the constraints on directly using internet-sourced lyrics due to copyright issues.  As a consequence, publicly available lyric datasets often resort to indirect methods of providing access to lyrics. These methods include offering code~\cite{kpop} or instructions~\cite{buffa2021wasabi} for web scraping commercial lyrics. An alternative strategy involves presenting a Bag-of-Words format, which includes the vocabulary of the lyrics along with its frequency but not the complete text. A notable example is the musiXmatch dataset~\cite{msd}, which provides the Bag-of-Words for lyrics of 237,662 songs. Despite lacking the full lyrical content, this dataset has been significantly instrumental owing to its alignment with the Million Song Dataset (MSD)~\cite{msd}, as the Bag-of-Words are paired with extensive metadata from various subsets of the MSD~\cite{schindler2012facilitating, korzeniowski2020mood} or other datasets correlated with the MSD's metadata~\cite{deezer}. Therefore, it has enhanced various lyric-based studies, including genre classification~\cite{liang2011music, yang2018lyric, wadhwa2021music}, emotion recognition~\cite{nanayakkara2016music}, cover song detection~\cite{correya2018large}, and psychology-based feature extraction~\cite{kim2020butter}. While the alignment with extensive metadata is beneficial, the absence of complete lyric content limits its applicability in areas that demand full text analysis, such as lyric generation, structural analysis, or in-depth lyric examination. For instance, while extracting unigrams using Bag-of-Words is feasible, generating $n$-grams (for $n$ > 1) remains unattainable, although we have identified several lyric studies that require $n$-gram-based analysis (for $n$ > 1)~\cite{watanabe2023text, ismir2023}. 

To address this challenge, we introduce a straightforward yet effective approach for reconstructing lyrics from the Bag-of-Words format, utilizing large language models. This method mitigates copyright concerns for academic use while preserving the substantial benefits derived from the extensive metadata associated with the original tracks. In the subsequent sections, we detail our lyric reconstruction methodology, present our dataset, LyCon, compiled using this approach and made accessible for academic purposes, highlight the potential applications of our suggested methods and dataset.

\begin{figure}
 \centerline{\framebox{
 \includegraphics[width=0.60\columnwidth]{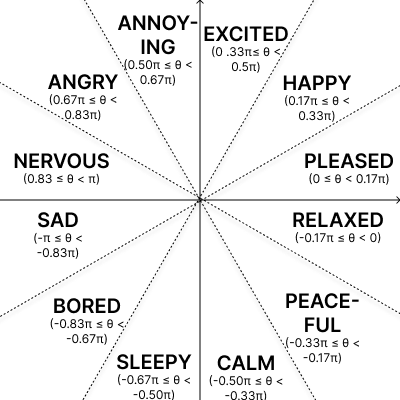}}}
 \caption{Table illustrating the relationship between $\theta$ values and [MOOD] categories. The valence levels (x-axis) and arousal levels (y-axis) determine $\theta$.}
 \label{fig:example}
\vspace{-3mm}
\end{figure}

\begin{figure}[h]
 \centerline{\framebox{
 \includegraphics[width=0.65\columnwidth]{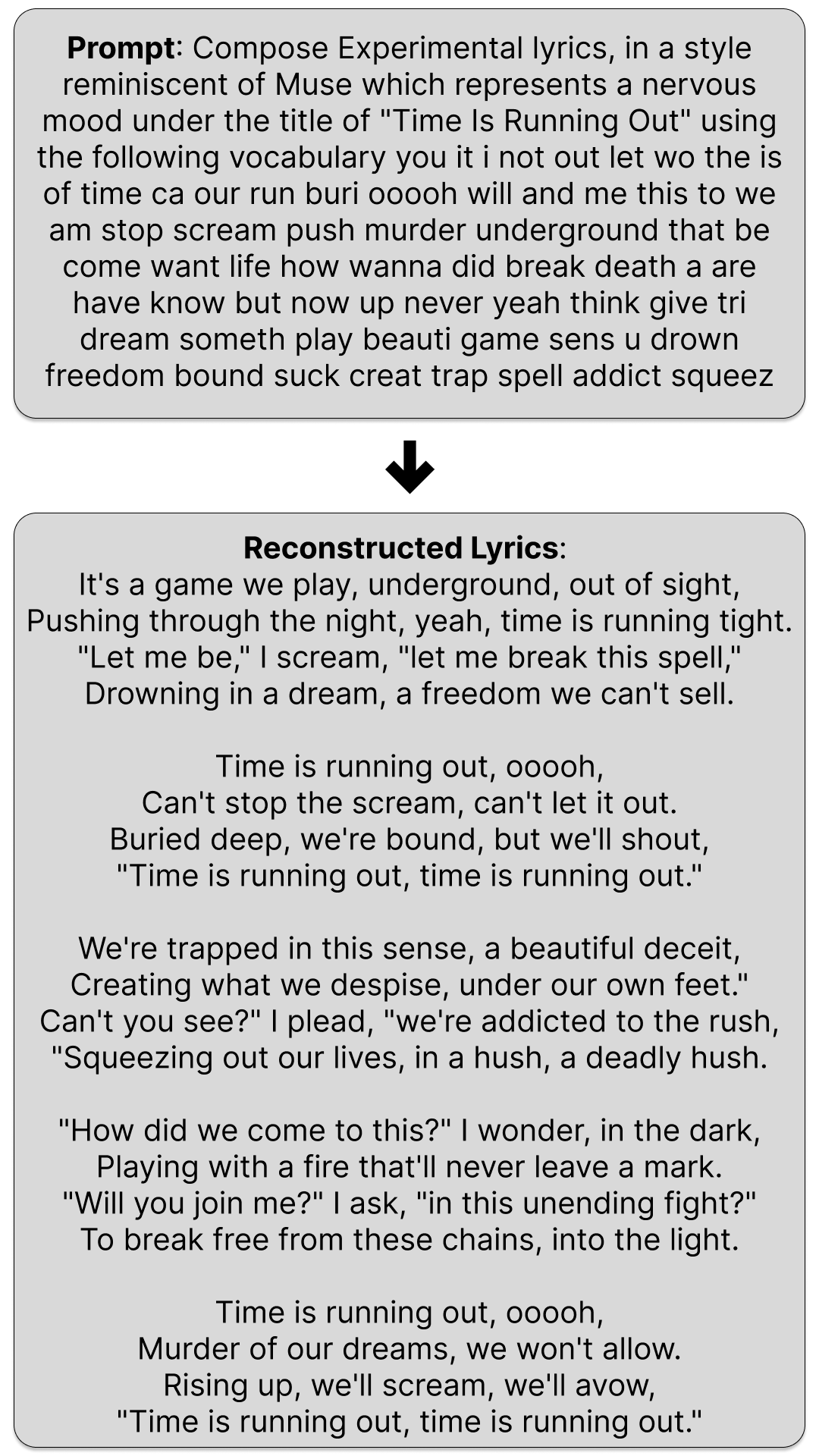}}}
 \caption{'Time Is Running Out' by Muse (MSD Track ID: TRSEKGD128F42B654D) as a prompting example, classified as the Experimental genre~\cite{schindler2012facilitating}, with the annotated emotional attributes of Valence -1.05 and Arousal 0.34~\cite{deezer} (thus, $\theta=0.90\pi$).}
 \label{fig:tiro}
 \vspace{-4mm}
\end{figure}

\section{Reconstruction Methodology}
Our primary objective is to reconstruct lyrics using large language models in a way that preserves their original vocabulary, topic, stylistic nuances, and mood. We accessed the vocabulary list for each song from the Bag-of-Words provided by the musiXmatch dataset, a subset of the MSD~\cite{msd}. To indirectly infer the topic, we used the title and artist information corresponding to each track. This approach is based on the premise that an artist's identity influences not just the style but also the thematic content of a song~\cite{tsukuda2017lyric}. Recognizing the correlation between a song's genre and its lyrical style, as extensively suggested in previous studies, we obtained the genre for each track from the Allmusic Genre Dataset~\cite{schindler2012facilitating}, another subset of the MSD. For mood annotations, we utilized the Deezer Mood Detection Dataset~\cite{deezer}, which offers emotional valence and arousal levels for musical tracks with its metadata aligned with the MSD. Utilizing this collected data, we direct a large language model~\footnote{We used the OpenAI's ChatGPT-4o.} to generate lyrics that fits the specific genre, reflect the mood, emulate the style of the corresponding artist under the song's title and employ the given Bag-of-Words, by prompting as below.
\begin{quote}
    
``Compose [GENRE] lyrics, in a style reminiscent of [ARTIST] which represents a [MOOD] mood under the title of [TITLE] using the following vocabulary [VOCABULARY].''
\end{quote}

In this study, the [GENRE] is determined by concatenating the genres listed in the Allmusic Style Dataset. The [ARTIST] and [TITLE] are directly taken as the artist's name and track title, respectively, from the MSD. The [VOCABULARY] is defined as a concatenation of vocabulary list from the Bag-of-Words, sorted in descending order of frequency. For [MOOD], we utilize the valence and arousal values from the Deezer Mood Detection Dataset. Following the 2D valence-arousal emotion space theory~\cite{russell1980circumplex}, which has been adopted in music domain~\cite{yang2012machine}, we represent valence as $x$ and arousal as $y$. The [MOOD] is then defined based on the angle in radians ($\theta$) between the positive $x$-axis and the point ($x$, $y$). The correlation between the range of $\theta$ values and their respective [MOOD] categories is illustrated in Figure 1. The example of the prompt and output is provided in Figure 2.

Using this methodology, we reconstruct lyrics against every songs that have associated metadata from all of the three datasets~\cite{msd, schindler2012facilitating, deezer} exploited for our prompt. This comprehensive list resulted in reconstructed lyrics for 7,863 songs. The reconstructed lyrics are downlodable at \url{https://github.com/havenpersona/lycon} where each item is mapped to the MSD song ID. 

\section{Analysis}
In Table~\ref{tab:example}, we compare the statistics of the reconstructed lyrics with those of the original lyrics for the selected songs. They exhibit statistical similarities in terms of average word, line, and section count, as well as the total count of unique bigrams and trigrams. Here, a gram is defined as a single word after simple whitespace separation, without considering capitalization. However, the total count of unique unigrams in LyCon is significantly lower than in the original lyrics, indicating a limited usage of diverse words. Finally, we compare the abstract and concrete words ratios, as defined in ~\cite{kao-jurafsky-2012-computational}, which suggests that aesthetically pleasing poetry is more likely to use concrete words over abstract words. As expected, LyCon shows a slightly higher abstract words ratio and a slightly lower concrete words ratio than the original lyrics. However, the small gap indicates that LyCon achieved an lesser yet similar aesthetic level to the original lyrics.

\section{Conclusions}
In this study, we presented a straightforward yet novel approach for reconstructing lyrics using the publicly available Bag-of-Words. This method offers a viable solution for research requiring complete lyrics without risking copyright infringement. Moreover, our publicly accessible compilation of reconstructed lyrics, enriched with metadata from three prominent datasets, is poised to advance lyric-based academic research, including mood and genre-conditioned lyric generation and music categorization based on lyrics.

\begin{table}[]
\centering
\resizebox{0.35\textwidth}{!}{%
\begin{tabular}{@{}lll@{}}
\toprule
\multicolumn{1}{c}{\textbf{Item}} & \multicolumn{1}{c}{\textbf{LyCon}} & \multicolumn{1}{c}{\textbf{Original}} \\ \midrule
Total Count of Lyrics Sets        & 7,863                              & 7,863                                 \\
Average Word Count per Set        & 319.42                             & 248.42                                \\
Average Line Count per Set        & 42.58                              & 38.62                                 \\
Average Section Count per Set     & 9.97                               & 9.86                                  \\
Total Count of Unique Unigrams    & 18,921                            & 36,222                                \\
Total Count of Unique Bigrams     & 269,139                            & 287,049                               \\
Total Count of Unique Trigrams    & 689,852                            & 593,959                               \\
Abstract Words Ratio              & 3.37                               & 2.82                                  \\
Concrete Words Ratio              & 2.92                               & 3.12                                  \\ \bottomrule
\end{tabular}
}
\caption{Statistical comparison between the reconstructed lyrics (LyCon) and original lyrics for the same list of songs.}
\label{tab:example}
\vspace{-3mm}
\end{table}

\bibliography{template}

%
%
%
%
%

\end{document}